\pdfoutput=1
\pdftrailerid{}
\documentclass[letterpaper]{article} \usepackage{aaai2026}  \usepackage{times}  \usepackage{helvet}  \usepackage{courier}  \usepackage[hyphens]{url}  \usepackage{graphicx}    \usepackage{natbib}  \usepackage{caption}     

\usepackage[utf8]{inputenc} \usepackage{booktabs}       \usepackage{amsfonts}              \usepackage{microtype}      \usepackage{xcolor}         \usepackage{amsmath}
\usepackage{amsthm}
\usepackage{amssymb}
\usepackage{subcaption}

\newtheorem{proposition}{Proposition}

\title{Learning the Latent Structure: A Feature-Centric Approach to Graph Data Augmentation}
\author {
Yu Song\textsuperscript{\rm 1},
    Zhigang Hua\textsuperscript{\rm 2},
    Yan Xie\textsuperscript{\rm 2},
    Bingheng Li\textsuperscript{\rm 1},
    Jingzhe Liu\textsuperscript{\rm 1},
    Bo Long\textsuperscript{\rm 2},
    Jiliang Tang\textsuperscript{\rm 1},
    Hui Liu\textsuperscript{\rm 1}
}
\affiliations {
\textsuperscript{\rm 1}Michigan State University\\
    \textsuperscript{\rm 2}Meta\\
    songyu5@msu.edu,
    zhua@meta.com,
    yanxie@meta.com,
    libinghe@msu.edu,
    liujin33@msu.edu,
    bolong@meta.com,
    tangjili@msu.edu,
    liuhui7@msu.edu
}

\begin{document}

\maketitle

\begin{abstract}
Graph-structured data plays a pivotal role in modeling complex relationships. However, real-world graphs are often incomplete due to data collection and observational constraints, severely limiting the effectiveness of modern graph learning pipelines. While existing Graph Data Augmentation (GDA) methods attempt to refine graph structures for improved downstream performance, they are typically label-dependent, computationally expensive, and inherently \textit{transductive}, limiting their applicability in practical scenarios.

In this work, we present a novel \textit{feature-centric} graph data augmentation framework that bypasses explicit structure modeling by operating directly in the embedding space. Through a self-supervised inverse masking process, our method captures latent ties between observed and complete graphs, enabling recovery of unobserved structural signals through refined node representations. To enhance robustness under noisy and sparse supervision, we introduce a message regularizer and a bootstrap strategy for effective training and generalization.
Evaluated on ten graph datasets spanning multiple domains, our approach, \textbf{SelfAug}, consistently outperforms state-of-the-art methods in both accuracy and efficiency across \textit{inductive} and \textit{cold-start} settings, highlighting its potential as a scalable and generalizable solution for real-world graph learning scenarios.
\end{abstract}

\section{Introduction}
Graphs offer a powerful framework for capturing intricate relationships. The inherent structure of graphs provides a valuable inductive bias about the non-independent and identically distributed (non-i.i.d.) nature of data, which can be effectively leveraged by graph learning methods such as Graph Neural Networks (GNNs)~\cite{kipf2016semi, velivckovic2017graph, chien2020adaptive} for improved performance on a variety of tasks from molecular structures to social networks~\cite{fan2019graph, rong2020self, zhou2020graph}. However, real-world graph data often suffers from incompleteness due to limitations in the data collection process and observational constraints~\cite{li2023gslb, zheng2021cold, kim2024content}. Consequently, many potential connections between entities remain unobserved, yielding sparse or suboptimal graph topologies that inadequately represent the true structure of the underlying system. In extreme scenarios, relational information may be completely unavailable, such as cold-start users in recommender systems~\cite{chen2020esam, kim2024content}.

Most existing graph learning models assume an ideal graph structure, neglecting its potential incompleteness~\cite{kipf2016semi, hou2022graphmae, yun2019graph}.
Dependence on such flawed topological information can degrade performance~\cite{li2023gslb}, and in severely sparse situations, even advanced GNNs may deteriorate to simple Multi-Layer Perceptrons (MLPs)~\cite{zheng2021cold}.
These limitations have motivated the development of Graph Data Augmentation (GDA)~\cite{ding2022data}, which aims to improve the quality of graph data by modifying or enriching the initial structure. Among these methods, some dynamically refine or infer graph connectivity during training, adapting to downstream applications in a task-specific manner ~\cite{fatemi2021slaps, franceschi2019learning, chen2020iterative, yu2021graph}. Others focus on test-time augmentation, altering the graph at inference to improve generalization or enhance robustness against adversarial perturbations~\cite{jin2022empowering, ju2023graphpatcher}.

Despite their effectiveness, existing GDA methods face critical limitations that hinder their practical applicability. First, most approaches rely heavily on supervision, requiring labeled data to guide the learning of improved graph structures~\cite{franceschi2019learning, chen2020iterative, jin2020graph, yu2021graph, fatemi2021slaps, ju2023graphpatcher, zhao2023graphglow}, which are often scarce or unavailable in realistic scenarios. Second, most methods exhibit poor generalization, as they learn the refined structure for a specific graph in a \textit{transductive} way, rendering them impractical for deployment on unseen or  dynamically evolving environments.
Third, efficiency remains a major bottleneck: many structure learning methods involve resource-intensive augmentation models and sophisticated optimization procedures such as bilevel optimization or iterative training~\cite{franceschi2019learning, chen2020iterative}. In addition, approaches that assess pairwise relationships between all nodes scale quadratically with the graph size~\cite{franceschi2019learning, liu2022towards, jin2022empowering}, making them unsuitable for large-scale applications.

To mitigate these limitations, we introduce a novel \textit{feature-centric} perspective for graph data augmentation, shifting the focus from explicit structure modeling to direct node representation enhancement. Rather than constructing or refining graph structure, our approach operates entirely in the embedding space, learning to adjust node representations in a way that reflects the influence of a more complete and informative structure, which fundamentally simplifies the augmentation process.
To circumvent the reliance on labeled data, we employ a self-supervised objective that learns the latent ties between observed and underlying structural information through an inverse masking process. This approach reconstructs complete representations from intentionally masked graphs, enabling the model to compensate for missing connections without requiring task-specific labels. Once trained, the augmenter operates on test graphs with a single forward pass to produce enriched node embeddings, avoiding the need for any dataset-specific optimization. Our method is scalable, label-free, and generalizable, making it well suited for real-world deployment. Our key contributions are summarized as follows:
\begin{itemize}
     \item We propose a novel \textit{feature-centric} perspective for graph data augmentation, shifting from traditional structure modeling to direct node representation enhancement. This new perspective simplifies the augmentation process and enables a more flexible and generalizable learning paradigm.

    \item We develop a self-supervised framework \textbf{SelfAug} that leverages intrinsic patterns of neighborhood formation to guide data augmentation.

    \item We demonstrate strong performance across inductive and cold-start settings, with significantly reduced inference cost compared to existing GDA baselines.
\end{itemize}

\section{Related Works}
\textbf{Graph Data Augmentation (GDA).} GDA encompasses a variety of methods aimed at enhancing downstream task performance by modifying graph data~\cite{ding2022data}. A prominent approach within GDA is graph structure learning, which concurrently learns an optimized structure for a given task and a task-specific solver such as a classifier for node classification~\cite{fatemi2021slaps, franceschi2019learning, chen2020iterative, yu2021graph, zhao2023graphglow, liu2022towards, chen2020learning, liu2021graph}. More recently, test-time adaptation techniques have been introduced to modify graphs to fit pre-trained GNN models, enabling infrastructure reuse without expensive retraining~\cite{jin2022empowering, ju2023graphpatcher}. However, these methods often rely on supervised signals and complex optimization routines, making them less scalable and poorly suited for generalization to unseen graphs.

\noindent\textbf{Graph Self-supervised Learning (GSSL).} To address label scarcity in real-world scenarios, self-supervised learning methods have gained significant traction. These methods largely fall into two categories: graph contrastive learning and graph generative modeling~\cite{hou2022graphmae, hou2023graphmae2, velivckovic2018deep, zhu2020deep, thakoor2021bootstrapped, you2020graph, song2024pure}. Graph SSL typically produces a graph encoder that generates embeddings directly useful for downstream tasks. While our approach can be viewed as a form of self-supervised learning (SSL) on graphs, it differs fundamentally from existing graph SSL methods in two key aspects: (1) it is explicitly motivated from a data augmentation perspective with tailored architectural designs and training objectives, whereas existing methods often overlook the imperfections in the observed graph; and (2) it is designed specifically for improved generalization capabilities, an aspect that is largely underexplored in current SSL approaches.

\section{Method}
\begin{figure*}[t]
  \centering
  \includegraphics[width=0.8\textwidth]{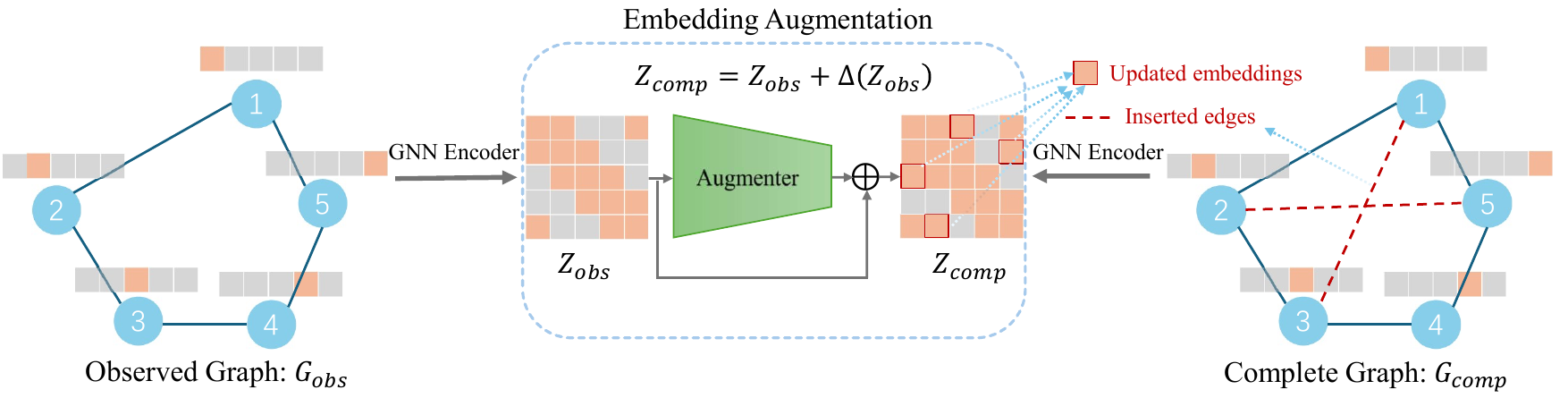}
  \caption{An overview of the proposed framework. \textbf{Left:} the observed graph $G_{\text{obs}}$ with missing edges. \textbf{Right:} the complete graph $G_{\text{comp}}$ with two inserted edges (red dashed lines). The GNN encoder transforms node features into embeddings that reflect each node’s 1-hop neighborhood. Our augmenter operates in the embedding space by predicting the residual between the observed embeddings $Z_{\text{obs}}$ and the complete embeddings $Z_{\text{comp}}$, thereby approximating the effect of a more complete graph structure without explicitly modifying the graph connectivity.}

  \label{fig:overview}
\end{figure*}

\subsection{Preliminaries}

Let $G = (A, X)$ denote a graph, where $A \in \{0, 1\}^{n \times n}$ represents the adjacency matrix encoding pairwise connections between $n$ nodes, and $X \in \mathbb{R}^{n \times d}$ denotes the node feature matrix with $d$-dimensional features for each node. Given a graph $G$, our goal is to predict target variables $Y$ (e.g., node labels in node classification tasks) through a function $f$ that maps the graph to the desired output space.
In this work, we focus on the problem of incomplete connections in the adjacency matrix $A$.
To tackle this issue, existing GDA methods~\cite{ding2022data, li2023gslb} aim to learn a modified graph $G' = (A', X')$ by transforming the original graph $G$. Generally, they can be defined as the following optimization problem:

\begin{equation} \label{eq:gda_general_reg_output}
\min_{\phi,\, \theta} \; \mathcal{L}(f_{\theta}(g_{\phi}(G)), Y) + \lambda \, \mathcal{R}(g_{\phi}(G))
\end{equation}

Here, $g_{\phi}$ denotes the graph augmenter, and $f_{\theta}$ is the prediction model (e.g., a GNN). $\mathcal{L}$ is the task-specific loss function (e.g., cross-entropy), and $\mathcal{R}$ is a regularization term applied to the augmented graph encouraging desirable properties such as sparsity.

Most existing GDA methods focus on modifying the graph structure \( A\)~\cite{fatemi2021slaps, franceschi2019learning, chen2020iterative, yu2021graph, zhao2023graphglow, liu2022towards}, although some also alter node features \( X\) to recover missing information or improve adversarial robustness~\cite{chen2020learning, liu2021graph, jin2020graph}. While effective, these approaches typically rely on heavily parameterized augmentation models and require complex optimization procedures. Furthermore, they inherently assume a \textit{transductive} setting, where the full test graph is accessible and optimization can be performed directly on it. These assumptions limit their practicality in real-world scenarios, where new graphs arrive continuously and cannot be optimized individually.

These challenges highlight the need for a lightweight graph augmenter that operates effectively in the \textit{inductive} setting. To this end, our key insight is that the influence of the graph structure is ultimately captured in the node embeddings through the neighborhood aggregation mechanism of graph learners such as GNNs. This suggests a promising alternative: rather than explicitly modeling or reconstructing the adjacency matrix, \textit{can we directly model the changes in node embeddings that would result from a more complete graph structure?}

Formally, let $Z_{obs} = f(A_{obs}, X)$ represent the node embeddings obtained from the observed (incomplete) graph using a graph encoder $f$, and $Z_{comp} = f(A_{comp}, X)$ represent the embeddings from a hypothetical complete graph with adjacency matrix $A_{comp}$. The difference between these embeddings can be captured by a node-wise modification function $\Delta$:

\begin{equation}
Z_{comp} = Z_{obs} + \Delta(Z_{obs})
\end{equation}

This formulation transforms the complex problem of structure learning into a node-wise learning task,
which fundamentally reduces the computational complexity from \( \mathcal{O}(n^2) \) to \( \mathcal{O}(n) \). The core intuition is that the augmenter \( \Delta \) learns to encode the latent relationship between the observed embeddings \( Z_{\text{obs}} \) and the embeddings derived from a more complete graph \( Z_{\text{comp}} \). In doing so, it captures the functional impact of missing connections on node representations, effectively learning to recover the information that would have been captured had those connections been observed.
Importantly, \( \Delta \) can be parameter-shared across nodes, making it inherently inductive: once trained, it can work on the embedding space of any graph from similar distributions without retraining, eliminating the need for graph-specific modeling such as parameterized adjacency matrices.
Next, we present the concrete design of our method, \textbf{SelfAug}, which implements the proposed idea in a principled and effective way. An overview is provided in Figure~\ref{fig:overview}.

\subsection{General Framework}

Building upon our preliminary insights, we now formalize our feature-centric graph data augmentation framework. At the core of our approach is an augmentation module \( \Delta_{\phi} \), parameterized by $\phi$, which operates on node embeddings derived from an observed graph. Given a graph encoder $f_{\theta}$, the augmenter outputs residual modifications that aim to compensate for the loss of structural information:
\begin{equation}
Z_{\text{aug}} = f_{\theta}(A_{\text{obs}}, X) + \Delta_{\phi}(f_{\theta}(A_{\text{obs}}, X))
\end{equation}

To train the augmenter \( \Delta_{\phi} \) without relying on downstream task labels, we propose a self-supervised framework with an inverse masking process. Specifically, we randomly drop a subset of edges from the observed graph to construct a synthetically sparser version, denoted as \( G_{\text{mask}} = (A_{\text{mask}}, X) \) where \( A_{\text{mask}} = A_{obs} \odot (1 - M) \), with \(M \in \{0,1\}^{n \times n} \) being a binary edge mask sampled from some distribution \( \mathcal{M} \).
Given the masked input, the model is trained to reconstruct the node embeddings computed from the unmasked (original) graph by minimizing the following reconstruction loss:

\begin{equation}
\label{eq:loss_1}
\begin{aligned}
\mathcal{L}_{\text{recon}} &= \mathbb{E}_{M \sim \mathcal{M}} \left[ d(\text{sg}(Z_{\text{obs}}), Z_{\text{mask}} + \Delta_{\phi}(Z_{\text{mask}})) \right], \\
Z_{\text{obs}} &= f_{\theta}(A_{\text{obs}}, X), \quad
Z_{\text{mask}} = f_{\theta}(A_{\text{mask}}, X)
\end{aligned}
\end{equation}

Here, \( d(\cdot, \cdot) \) is a distance function, and \(\text{sg}\) denotes the stop-gradient operation. In practice, we define \( d(\cdot, \cdot) \) as a weighted combination of cosine distance and mean squared error to jointly capture semantic alignment and scale information:
\[
d(x, y) = \lambda_{\text{cos}} \cdot \left(1 - \cos(x, y)\right) + \lambda_{\text{mse}} \cdot \|x - y\|_2^2,
\]
where \(\lambda_{\text{cos}}\) and \(\lambda_{\text{mse}}\) are hyperparameters that control the trade-off between directional similarity and magnitude consistency.

Below, we provide theoretical justification showing that this self-supervised learning objective effectively guides the augmenter to enrich node representations with information beneficial for downstream prediction tasks. Due to space limit, we leave the detailed proof in Appendix 1.

\begin{proposition}
Let \( A_{obs} \in \{0,1\}^{n \times n} \) be the adjacency matrix of an observed graph and \( M \in \{0,1\}^{n \times n} \) be a binary edge mask sampled from some distribution \( \mathcal{M} \). Define the masked adjacency matrix as \( A_{\text{mask}} = A_{obs} \odot (1 - M) \), where \( \odot \) denotes element-wise multiplication. Let \( X \in \mathbb{R}^{n \times d} \) be the node feature matrix and \( Y \) be the target labels for some prediction task. The mutual information between $Y$ and $(A_{mask}, X)$ is bounded as:
\[
I(Y; A_{\text{mask}} , X) \leq I(Y; A_{obs} , X)
\]
\end{proposition}

\textbf{Remark.}
Proposition 1 implies that masking edges leads to a loss of mutual information between the input graph and the target labels, i.e., the masked graph \( A_{\text{mask}} \) is less informative than the original graph \( A_{obs} \) for downstream prediction. This justifies the design of our reconstruction objective in Eq.~\ref{eq:loss_1}, which seeks to compensate for this information loss by learning a correction \( \Delta_\phi \) that restores the embeddings \( Z_{\text{mask}} \) toward their more informative counterpart \( Z_{\text{obs}} \). As a result, the model is encouraged to infer and recover the missing structural signal in a way that preserves task-relevant information, even without direct label supervision.

We now instantiate the design of the augmenter \( \Delta_{\phi} \), which transforms node embeddings from a masked graph into enriched embeddings that reflect complete structural information. In the following, we show that a simple multi-layer perceptron (MLP) is sufficient to approximate the embedding differences between masked and unmasked views, and can be effectively optimized via gradient-based training.

\begin{proposition}
Let $f_\theta$ be a fixed GNN encoder with parameters $\theta$ and define the embeddings
\begin{equation}
\begin{aligned}
Z_{\mathrm{obs}} &= f_\theta(A_{\mathrm{obs}}, X) \in \mathbb{R}^{n \times h}, \\
Z_{\mathrm{mask}} &= f_\theta(A_{\mathrm{mask}}, X) \in \mathbb{R}^{n \times h}
\end{aligned}
\label{eq:shared_encoding}
\end{equation}

where $X$, $A_{obs}$ and $A_{mask}$ are defined in the same way as in Proposition 1.
Let $\Delta_\phi:\mathbb{R}^{n\times h}\!\to\!\mathbb{R}^{n\times h}$ be a fully-connected MLP with ReLU activations and sufficient width and depth. Then the following holds:

\begin{enumerate}
\item \textbf{Expressivity.}
      For any $\varepsilon>0$ there exists a choice of parameters~$\phi$ such that
      \(
      \bigl\|\Delta_\phi(Z_{\mathrm{mask}})\!-\!(Z_{\mathrm{obs}}\!-\!Z_{\mathrm{mask}})\bigr\|_F<\varepsilon
      \)
      for all masks $M\!\sim\!\mathcal{M}$.  Hence the MLP can approximate the exact residual between masked and full-graph embeddings arbitrarily well.

\item \textbf{Optimizable Objective.}
      With a smooth distance metric $d$ (e.g., squared Frobenius norm), the reconstruction loss
      \[
      \mathcal{L}_{\mathrm{recon}}(\phi)=
      \mathbb{E}_{M\sim\mathcal{M}}
      \bigl[d\bigl(Z_{\mathrm{obs}},
                  Z_{\mathrm{mask}}+\Delta_\phi(Z_{\mathrm{mask}})\bigr)\bigr]
      \]
      is locally Lipschitz and almost everywhere differentiable in~$\phi$; thus first-order optimizers such as SGD converge to a stationary point.
\end{enumerate}
\end{proposition}

\textbf{Remark.}
Proposition 2 establishes that a suitable MLP can approximate the ideal correction \(Z_{\mathrm{obs}} - Z_{\mathrm{mask}}\) with arbitrary precision and that the associated reconstruction loss is well behaved for gradient-based optimization. Consequently, it is theoretically feasible to train the augmentation module to recover information lost due to edge masking. In practice, we adopt a lightweight three-layer MLP with a moderate hidden dimension (e.g., 384), which empirically proves effective across evaluated datasets.

It is worth noting that during training, we treat the observed graph \( A_{\text{obs}} \) as the latent, ideal structure and train the augmenter to reconstruct its effects from a masked counterpart. This approach relies on the assumption that the training graph is relatively \textit{clean} and \textit{dense}—\textit{clean} indicates that it reflects meaningful and generalizable connection patterns, and \textit{dense} in the sense that it provides sufficient structural supervision for the augmenter to learn from.
However, these two assumptions may not hold in reality. In the following subsections, we introduce two techniques to improve robustness under noisy and sparse training environments through a message regularizer and a bootstrap training strategy, respectively.

\subsection{Capturing Reliable Relationships}

Since the ideal graph \( A_{\text{comp}} \) is not accessible during training, we use the observed structure \( A_{\text{obs}} \) as a proxy. A key challenge, however, is that the training graph itself may be noisy or arbitrary. That is, connections may form due to incidental or domain-specific reasons, resulting in edges that do not reflect generalizable structural patterns. Directly learning from such graphs risks overfitting to spurious correlations and degrades the model’s ability to generalize.

Consider a citation network, where nodes represent papers and edges represent citations. While some citations reflect meaningful topical relevance (i.e., stable, transferable patterns), others may be perfunctory or irrelevant (i.e., noise). An effective augmentation mechanism should emphasize the former while de-emphasizing the latter to support generalization across diverse test graphs.

To this end, we introduce a message regularizer that encourages the model to focus on stable and reliable neighborhood information during aggregation. Given the self-supervised objective in Eq.~\ref{eq:loss_1}, our goal is to enable the encoder \( f_{\theta} \) to filter out unrepresentative signals passed between nodes. Concretely, we devise a regularizer that encourages the model to enforce similarity in message representations between sufficiently similar nodes.
Formally, let \( h_i^{(l-1)} \) be the embedding of node \( i \) at layer \( l-1 \) of the GNN encoder, and let \( m_i^{(l)} \) denote the message aggregated from its neighbors \( N_i \) at layer \( l \). We define the message regularization loss at layer \( l \) as:
\begin{equation} \label{eq:msg_reg_main}
\mathcal{L}_{\text{msg}}^{(l)} = \frac{1}{Z^{(l)}} \sum_{i \neq j} w_{ij}^{(l)} \cdot d_{\text{msg}}(m_i^{(l)}, m_j^{(l)})
\end{equation}
Here, \( d_{\text{msg}}(m_i^{(l)}, m_j^{(l)}) \) denotes the distance between the messages of nodes \( i \) and \( j \), and \( w_{ij}^{(l)} \) is a pairwise weighting function defined as:
\begin{equation} \label{eq:weight_func_main}
w_{ij}^{(l)} =
\begin{cases}
(s_{ij}^{(l-1)})^2 & \text{if } s_{ij}^{(l-1)} \geq \tau \text{ and } i \neq j \\
0 & \text{otherwise}
\end{cases}
\end{equation}
where \( s_{ij}^{(l-1)} = \text{cos}(h_i^{(l-1)}, h_j^{(l-1)}) \) is the cosine similarity between the embeddings of nodes \( i \) and \( j \) from the previous layer, and \( \tau \) is a similarity threshold hyperparameter. The squaring of similarity amplifies the regularization strength for more similar node pairs. The normalization factor \( Z^{(l)} = \sum_{i \neq j} \mathbb{I}(w_{ij}^{(l)} > 0) \) ensures scale invariance across layers.
In practice, we instantiate the message distance as cosine distance, \( d_{\text{msg}}(x, y) = 1 - \text{cos}(x, y) \).

With this message regularizer, the model is guided to focus on shared, robust neighborhood characteristics rather than noisy or idiosyncratic ones, reinforcing the learning of semantics that are common to similar nodes and thus more likely to generalize.

\subsection{Bootstrap Augmentation}
\label{sec:bootstrap}

We now address the challenge posed by the sparsity of the training graph. In particular, the training graph \( A \) may be incomplete or sparse, offering limited structural information and resulting in weak supervision signals for learning the augmenter \( \Delta_{\phi} \).

To mitigate this issue and enhance the training process, we introduce a bootstrap optimization strategy inspired by self-distillation~\cite{zhang2019your}. Rather than treating \( Z_{\text{obs}} \) as the fixed ground truth, we hypothesize that the augmenter—once it has captured meaningful structural patterns—can provide useful corrections to \( Z_{\text{obs}} \) itself, nudging it toward a richer representation of an idealized, more complete graph. Accordingly, we dynamically refine the target embeddings by incorporating the augmenter's predictions during training.

Formally, at each training epoch \( t \), we compute a bootstrapped target embedding \( Z_{\text{target}}^{(t)} \) as follows:
\begin{equation}
Z_{\text{target}}^{(t)} = Z_{\text{obs}} + \alpha^{(t)} \cdot \Delta_{\phi}(Z_{\text{obs}})
\label{eq:bootstrap_target}
\end{equation}
Here, \( \Delta_{\phi}(Z_{\text{obs}}) \) is the augmentation predicted when the input is the original (observed) graph, and \( \alpha^{(t)} \in [0,1] \) is a scheduling coefficient that controls the influence of the augmenter's prediction at epoch \( t \). In our implementation, we adopt a linearly increasing schedule for \( \alpha^{(t)} \), gradually shifting reliance from the static target \( Z_{\text{obs}} \) to the refined target as training progresses.

The self-supervised reconstruction loss is then updated to use the bootstrapped target (for clarity, we omit the epoch index \( t \) below):
\begin{equation}
\label{eq:loss_3}
\mathcal{L}_{\text{recon}} = \mathbb{E}_{M \sim \mathcal{M}} \left[ d(\text{sg}(Z_{\text{target}}), Z_{\text{mask}} + \Delta_{\phi}(Z_{\text{mask}})) \right]
\end{equation}

\section{Experiments}

In this section, we aim to answer a central question:
\textit{can \textbf{SelfAug} work effectively to enrich node representations and generalize to unseen test graphs?} To this end, we evaluate our method under challenging inductive and cold-start settings to demonstrate its effectiveness and efficiency, while also providing deeper insights into the behavior of the augmenter and the impact of each proposed module.

\subsection{Experiment Settings}

\begin{table*}[t]
\centering
\small \setlength{\tabcolsep}{1.5pt}{\renewcommand{\arraystretch}{0.8}\begin{tabular}{l c c c c c c c c c c c}
\toprule
& \shortstack{\textbf{Training}\\\textbf{Data}}
& \textsc{Cora}
& \textsc{Citeseer}
& \textsc{Pubmed}
& \textsc{DBLP}
& \textsc{WikiCS}
& \shortstack{\textsc{ogbn-}\\\textsc{arxiv}}
& \textsc{Sportsfit}
& \textsc{Products}
& \textsc{Photo}
& \textsc{Computer} \\
\midrule
\multicolumn{12}{c}{\textbf{Graph Neural Networks}} \\
\midrule
GCN & $X_\text{test}, A_\text{test}, Y_\text{test}$ & 72.30 & 65.66 & 68.41 & 59.83 & 58.71 & \underline{51.95} & 60.86 & 39.18 & 57.78 & 56.80 \\
GAT & $X_\text{test}, A_\text{test}, Y_\text{test}$ & 72.65 & 65.83 & 67.75 & 62.00 & 60.04 & 50.60 & 63.51 & 36.50 & 60.09 & 60.56 \\
APPNP & $X_\text{test}, A_\text{test}, Y_\text{test}$ & 72.60 & 65.23 & 67.38 & 61.58 & 59.51 & 49.91 & 61.22 & 38.66 & 57.60 & 58.57 \\
GPRGNN & $X_\text{test}, A_\text{test}, Y_\text{test}$ & 73.54 & 66.21 & 68.49 & 62.00 & 59.33 & 50.09 & 62.39 & 38.51 & 60.09 & 62.51 \\
\midrule
\multicolumn{12}{c}{\textbf{Graph Self-supervised Learning}} \\
\midrule
GraphMAE & $X_\text{train}, A_\text{train}$ & 74.06 & \underline{72.21} & 71.22 & 70.13 & 60.31 & 48.90 & 61.58 & \underline{40.96} & 65.02 & 61.90 \\
GraphMAE2 & $X_\text{train}, A_\text{train}$ & 72.17 & 68.17 & 72.24 & 67.04 & 58.97 & 48.51 & 59.27 & 39.18 & 61.15 & 58.79 \\
VGAE & $X_\text{train}, A_\text{train}$ & 74.06 & 69.62 & \underline{75.51} & 69.88 & 58.00 & 49.40 & \underline{63.69} & 38.85 & \underline{65.87} & \underline{66.74} \\
DGI & $X_\text{train}, A_\text{train}$ & \underline{76.13} & 69.02 & 71.92 & \underline{70.96} & 57.47 & \texttt{OOM} & 63.60 & 39.33 & 64.04 & 61.99 \\
GRACE & $X_\text{train}, A_\text{train}$ & 75.35 & 70.13 & 72.82 & 69.67 & \underline{61.87} & \texttt{OOM} & 63.60 & 38.42 & 64.22 & 66.48 \\
\midrule
\multicolumn{12}{c}{\textbf{Graph Data Augmentation}} \\
\midrule
GRCN & $X_\text{test}, A_\text{test}, Y_\text{test}$ & 73.20 & 62.17 & 63.92 & 64.38 & 53.56 & 48.92 & 63.65 & 38.45 & 59.65 & 59.07 \\
IDGL & $X_\text{test}, A_\text{test}, Y_\text{test}$ & 70.71 & 62.17 & 67.46 & 59.33 & 41.33 & \texttt{OOM} & 61.48 & 38.64 & 57.01 & 56.49 \\
ProGNN & $X_\text{test}, A_\text{test}, Y_\text{test}$ & 70.92 & 61.87 & 68.33 & 58.67 & 58.53 & \texttt{OOM} & 61.12 & 37.90 & 59.43 & 56.50 \\
SUBLIME & $X_\text{test}, A_\text{test}$ & 71.14 & 59.11 & 62.56 & 59.71 & 43.78 & \texttt{OOM} & 62.31 & 38.43 & 60.91 & 59.73 \\
\midrule
SelfAug (Ours) & $X_\text{train}, A_\text{train}$
& \textbf{77.37}
& \textbf{74.85}
& \textbf{77.53}
& \textbf{76.20}
& \textbf{68.80}
& \textbf{53.01}
& \textbf{65.81}
& \textbf{44.03}
& \textbf{71.60}
& \textbf{72.66} \\
\bottomrule
\end{tabular}}
\caption{Performance comparison for inductive node classification. Results are reported in terms of classification accuracy (\%). \textbf{Bold} numbers indicate the best performance per dataset, and \underline{underlined} numbers indicate the second-best. \texttt{OOM} denotes out-of-memory errors during training.}
\label{tab:performance_comparison_inductive}
\end{table*}

\begin{table*}[t]
\centering
\small \setlength{\tabcolsep}{1.5pt}{\renewcommand{\arraystretch}{0.8}\begin{tabular}{l c c c c c c c c c c c}
\toprule
& \shortstack{\textbf{Training}\\\textbf{Data}}
& \textsc{Cora}
& \textsc{Citeseer}
& \textsc{Pubmed}
& \textsc{DBLP}
& \textsc{WikiCS}
& \shortstack{\textsc{ogbn-}\\\textsc{arxiv}}
& \textsc{Sportsfit}
& \textsc{Products}
& \textsc{Photo}
& \textsc{Computer}\\
\midrule
MLP & $X_\text{test}, Y_\text{test}$ &
63.70 &
62.38 &
68.87 &
57.17 &
55.91 &
47.14 &
56.31 &
34.34 &
51.60 &
41.47 \\
\midrule
\multicolumn{12}{c}{\textbf{Graph Self-supervised Learning}} \\
\midrule
GraphMAE & $X_\text{train}, A_\text{train}$
& 68.34 & \underline{70.64} & 71.88 & 65.71 & \underline{61.73} & 45.17 & 61.22 & 38.03 & 62.44 & 58.75\\
GraphMAE2 & $X_\text{train}, A_\text{train}$
& 71.65 & 68.12 & 72.37 & 67.75 & 59.91 & \underline{48.64} & 57.56 & 37.36 & 61.64 & 51.36\\
VGAE & $X_\text{train}, A_\text{train}$ &
69.81 & 67.40 & \underline{76.41} & 63.58 & 58.89 & 45.38 & 62.57 & \textbf{39.47} & \underline{64.36} & 58.96\\
DGI & $X_\text{train}, A_\text{train}$ &
\underline{71.96} & 65.96 & 72.16 & \underline{68.21} & 55.51 & \texttt{OOM} & \underline{62.66} & 37.17 & 60.36 & 55.38\\
GRACE & $X_\text{train}, A_\text{train}$
& 68.56 & 69.83 & 72.62 & 63.75 & 56.62 & \texttt{OOM} & 61.44 & 37.51 & 63.16 & \underline{59.44}\\
\midrule
\multicolumn{12}{c}{\textbf{Graph Data Augmentation}} \\
\midrule
GRCN    & $X_\text{test}, Y_\text{test}$
        & 68.17
        & 59.79
        & 64.58
        & 60.54
        & 50.18
        & 47.24
        & 59.81
        & 36.40
        & 54.73
        & 48.94 \\
IDGL    & $X_\text{test}, Y_\text{test}$
        & 65.68
        & 61.74
        & 67.09
        & 59.50
        & 54.76
        & \texttt{OOM}
        & 58.73
        & 35.79
        & 54.05
        & 44.76\\
ProGNN  & $X_\text{test}, Y_\text{test}$
        & 63.83
        & 58.77
        & 68.12
        & 55.54
        & 54.93
        & \texttt{OOM}
        & 58.56
        & 36.82
        & 52.42
        & 44.49\\
SUBLIME & $X_\text{test}$
        & 60.13
        & 46.89
        & 62.06
        & 42.96
        & 33.02
        & \texttt{OOM}
        & 57.10
        & 36.45
        & 52.26
        & 49.19\\
\midrule
SelfAug (Ours) & $X_\text{train}, A_\text{train}$
& \textbf{75.31}
& \textbf{73.95}
& \textbf{76.62}
& \textbf{72.70}
& \textbf{70.53}
& \textbf{50.98}
& \textbf{64.86}
& \underline{38.51}
& \textbf{68.58}
& \textbf{62.12} \\
\bottomrule
\end{tabular}}
\caption{Performance comparison for cold-start node classification. Results are reported in terms of classification accuracy (\%). \textbf{Bold} numbers indicate the best performance per dataset, and \underline{underlined} numbers indicate the second-best. \texttt{OOM} denotes out-of-memory errors during training.}
\label{tab:performance_comparison_cold_start}
\end{table*}

\textbf{Datasets.} We evaluate our method on ten widely adopted graph benchmarks: \textsc{Cora}, \textsc{Citeseer}, \textsc{Pubmed}, \textsc{DBLP}, \textsc{WikiCS}, and \textsc{ogbn-arxiv}, which are academic citation networks or Wikipedia connectivity graphs; as well as \textsc{Sportsfit}, \textsc{Products}, \textsc{Photo}, and \textsc{Computer}, which are e-commerce networks derived from Amazon's interaction data. Notably, \textsc{ogbn-arxiv} is a large-scale benchmark comprising over 160,000 nodes, providing a testbed for scalability evaluation. To ensure consistent and fair comparisons, we utilize the preprocessed versions from~\cite{chen2024exploring}, where node features are derived from textual attributes using SentenceBERT embeddings~\cite{thakur-2020-AugSBERT}.

\noindent\textbf{Evaluation Protocols.} To assess performance in realistic scenarios, we introduce two evaluation settings: \textit{inductive} and \textit{cold-start} node classification, following~\cite{li2023gslb}. In the \textit{inductive}  setting, each dataset is split into three disjoint subgraphs: \( G_{\text{train}} \), \( G_{\text{val}} \), and \( G_{\text{test}} \), such that their node and edge sets are mutually exclusive (i.e., \( V_{\text{train}} \cap V_{\text{val}} = \emptyset \), \( E_{\text{train}} \cap E_{\text{val}} = \emptyset \), etc.). We employ random splits for all datasets except \textsc{ogbn-arxiv}, with which we adopt the original split based on paper publication dates, reflecting a more realistic evaluation scenario.
The \textit{cold-start} setting simulates a more challenging scenario in which all edges in \( G_{\text{val}} \) and \( G_{\text{test}} \) are removed, leaving only node features available for prediction. This setting reflects practical applications such as cold-start recommendation~\cite{chen2020esam}, where no relational information is available for new users or items.
For each validation/test graph, we randomly sample 5 labeled nodes per class for training, reserve a validation set for hyperparameter tuning, and use the remaining nodes for testing. Test accuracy is reported as the average over 5 random splits to ensure statistical robustness.
In both settings, our augmentation model is trained on \( G_{\text{train}} \) and evaluated on \( G_{\text{val}} \) and \( G_{\text{test}} \) without any retraining or fine-tuning, thereby assessing its ability to generalize to newly arriving graphs.
Detailed dataset statistics are provided in Appendix 2.

\noindent\textbf{Baselines.} We compare our method against representative baselines from three categories: (i) \textit{Advanced GNN Architectures}, including GCN~\cite{kipf2016semi}, GAT~\cite{velivckovic2017graph}, APPNP~\cite{gasteiger2018predict}, and GPRGNN~\cite{chien2020adaptive}, which serve as non-augmented model baselines; (ii) \textit{Graph Self-Supervised Learning (GSSL) Methods}, such as DGI~\cite{velivckovic2018deep}, GRACE~\cite{zhu2020deep}, GraphMAE~\cite{hou2022graphmae}, GraphMAE2~\cite{hou2023graphmae2} and VGAE~\cite{kipf2016variational}, which, similar to our method, are trained in a label-free manner on \( G_{\text{train}} \) and used to generate embeddings for \( G_{\text{val}} \) and \( G_{\text{test}} \); and (iii) \textit{Graph Data Augmentation (GDA) Methods}, including GRCN~\cite{yu2021graph}, IDGL~\cite{chen2020iterative}, ProGNN~\cite{jin2020graph} and SUBLIME~\cite{liu2022towards}.

\noindent\textbf{Implementation Details.} We adopt a two-layer Graph Attention Network (GAT) with a hidden dimension of 384 as our graph encoder.
To prevent model collapse, we maintain two copies of the encoder during training: a primary encoder \( f_{\theta} \), which processes masked graphs and is updated via gradient descent, and a target encoder \( f_{\theta'} \), which computes target embeddings from the unmasked graph, and is updated using an Exponential Moving Average (EMA), following prior work in bootstrapped self-supervision~\cite{grill2020bootstrap, thakoor2021bootstrapped}.
The augmentation module \( \Delta \) is implemented as a three-layer MLP, with a hidden dimension of 384. We train the model using the Adam optimizer.
For the SSL baselines, we choose either GCN or GAT as the backbone depending on validation performance.
For our method, as well as for GNN and graph SSL baselines, we apply Optuna for hyperparameter tuning with a search budget of 100 trials, optimizing learning rate, weight decay, and model-specific parameters. For GDA methods, we adopt the implementation and default settings from~\cite{li2023gslb}.
Additional implementation details are provided in Appendix 3.

\subsection{Performance Comparison}
\label{sec:comparison}

We present the results for inductive node classification in Table~\ref{tab:performance_comparison_inductive}, where ``Training Data'' indicates the data available during training. GNN methods utilize the test graph with \( (X_{\text{test}}, A_{\text{test}}, Y_{\text{test}}) \) in the traditional semi-supervised setting. In contrast, SSL methods learn a graph encoder solely from \( (X_{\text{train}}, A_{\text{train}}) \), and transfer it to unseen test graphs to compute node representations. GDA methods operate directly on \( G_{\text{test}} \), and typically rely on \( Y_{\text{test}} \) to guide graph refinement, with SUBLIME being an exception that performs structure refinement in a self-supervised manner.

As shown in Table~\ref{tab:performance_comparison_inductive}, our method \textbf{SelfAug} achieves the highest accuracy across all datasets. Notably, it outperforms the second best method by a considerable margin, e.g., 5.24\% on \textsc{DBLP} and 5.73\% on \textsc{Photo}. Importantly, \textbf{SelfAug} is trained without access to the test graph or its labels, underscoring its strong generalization capability in inductive settings.
In contrast, GDA methods perform unstably and struggle to surpass the base GNNs. This is primarily due to their reliance on scarce supervision and the complexity of graph structure optimization, making them less robust for real-world applications.

We observe that SSL methods generally outperform supervised GNNs, thanks to their ability to exploit unlabeled data. However, their performance lags on the \textsc{ogbn-arxiv} dataset, which uses a timeline-aware split based on publication dates. This temporal split introduces more pronounced distributional shifts compared to the random splits used in other datasets, making it challenging for existing SSL methods to generalize to the test graph, even though they originate from the same domain.
By comparison, \textbf{SelfAug} consistently outperforms both supervised and SSL baselines. This arises from its carefully designed training objectives, which effectively leverages unlabeled data to capture the underlying distribution of neighboring features, along with its specific focus on generalizable pattern discovery, making it particularly suited for inductive scenarios. This advantage is further validated in the cold-start setting (Table~\ref{tab:performance_comparison_cold_start}), where similar trends are observed.

\subsection{Augmenter Analysis}
\begin{figure}[t]
  \centering
  \begin{subfigure}[t]{0.49\linewidth}
    \centering
    \includegraphics[width=\linewidth]{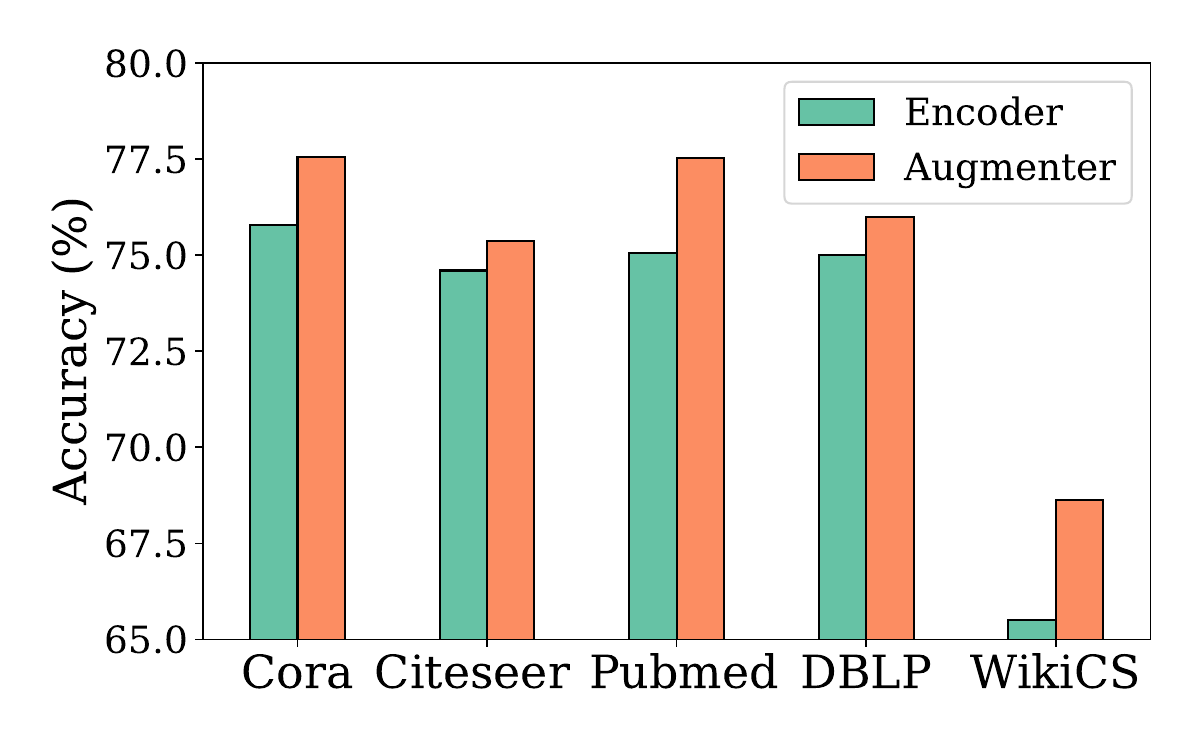}
    \caption{Test Accuracy}
    \label{fig:acc_half}
  \end{subfigure}\hfill
  \begin{subfigure}[t]{0.49\linewidth}
    \centering
    \includegraphics[width=\linewidth]{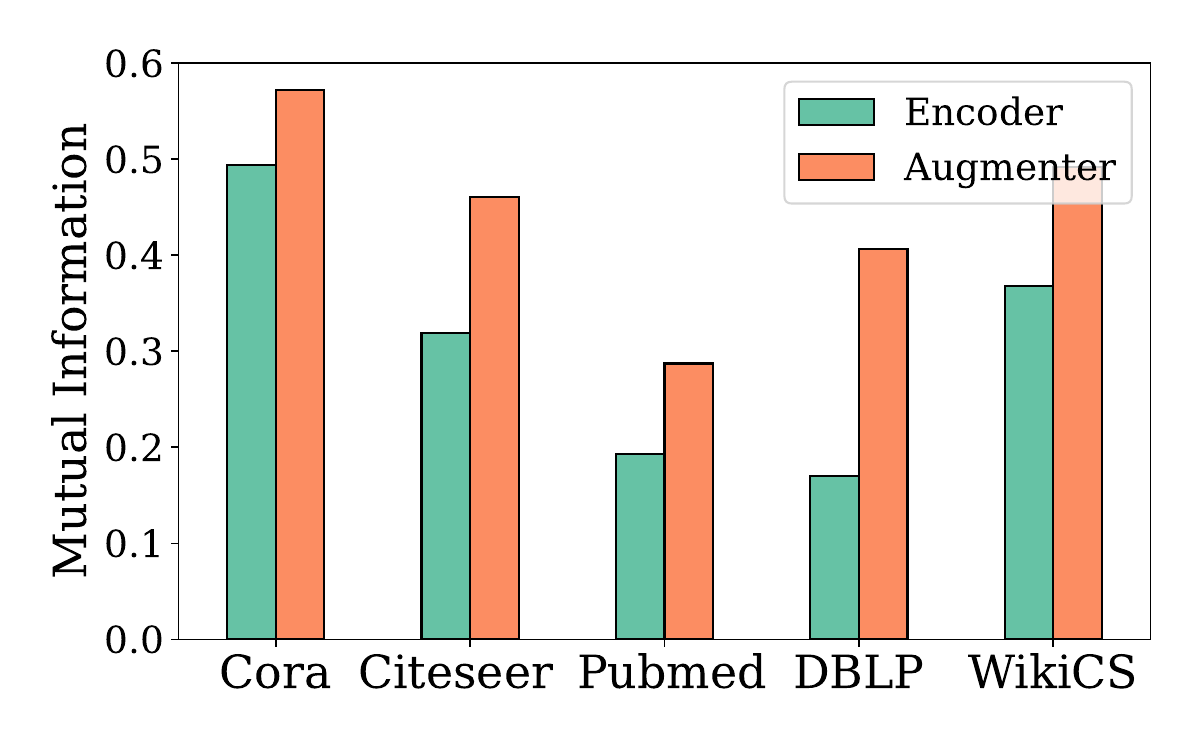}
    \caption{Mutual Information}
    \label{fig:mi_half}
  \end{subfigure}

  \caption{Comparison of embeddings before and after going through the augmentation model, denoted as `Encoder' and `Augmenter', respectively. We report the test accuracy and mutual information between embeddings and labels.}
  \label{fig:discussion}
\end{figure}

To better understand the effect of our augmentation module \( \Delta_{\phi} \), we compare the quality of node embeddings before and after going through the augmenter. Specifically, we contrast the raw GNN encoder output \(Z_{enc} = f_{\theta}(A, X) \) with the enriched embeddings \( Z_{\text{aug}} = f_{\theta}(A, X) + \Delta_{\phi}(f_{\theta}(A, X)) \).
Figure~\ref{fig:discussion} reports results on five datasets using two metrics: (a) test accuracy with a linear classifier, and (b) mutual information (MI), computed as normalized mutual information (NMI) between k-means clustering and ground-truth labels.

We observe consistent improvements in both metrics, confirming that the augmenter effectively improves embedding quality for the node classification task. Notably, the MI gain is often larger than the accuracy gain, likely because MI more directly measures alignment between embeddings and label structure (while accuracy involves an additionally trained linear classifier which may compensate for the original embeddings). These results further validate that our augmenter captures generalizable structural patterns and improves embedding quality in a self-supervised manner.

\subsection{Efficiency Analysis}

As shown in Table~\ref{tab:efficiency}, \textbf{SelfAug} achieves the best overall runtime and memory usage, which is comparable to a simple 2-layer GCN. This efficiency arises from its lightweight inference process: it performs only a forward pass on the test graph without any retraining or fine-tuning, and uses a linear classifier on the resulting embeddings for classification.
This design makes it highly suitable for fast and resource-efficient deployment. By contrast, structure learning methods rely on heavier models and require dataset-specific optimization, resulting in significantly higher computational overhead.

\begin{table}[t]
\centering
\small
\setlength{\tabcolsep}{2pt}{\renewcommand{\arraystretch}{0.75}\begin{tabular}{lcc}
\toprule
\textbf{Method} & \textbf{Ave Time (s)} & \textbf{Ave Mem (MB)} \\
\midrule
GCN     & \textbf{1.46}\ddag & \textbf{15.31}\dag \\
GAT     & 2.05               & 79.98              \\
\midrule
GRCN    & 6.23               & 78.00              \\
IDGL    & 6.09               & 476.50             \\
ProGNN  & 69.03              & 78.85              \\
SUBLIME & 92.27              & 133.03             \\
\midrule
SelfAug & \textbf{0.90}\dag  & \textbf{15.83}\ddag \\
\bottomrule
\end{tabular}}
\caption{Comparison of average runtime till convergence (in seconds) and peak GPU memory usage (in MB) across methods on \textsc{Cora}. \textbf{\dag} and \textbf{\ddag} denote the best and second-best results, respectively.}
\label{tab:efficiency}
\end{table}

\subsection{Ablation Study}
\begin{figure}[t]
\centering
\includegraphics[width=0.75\linewidth]{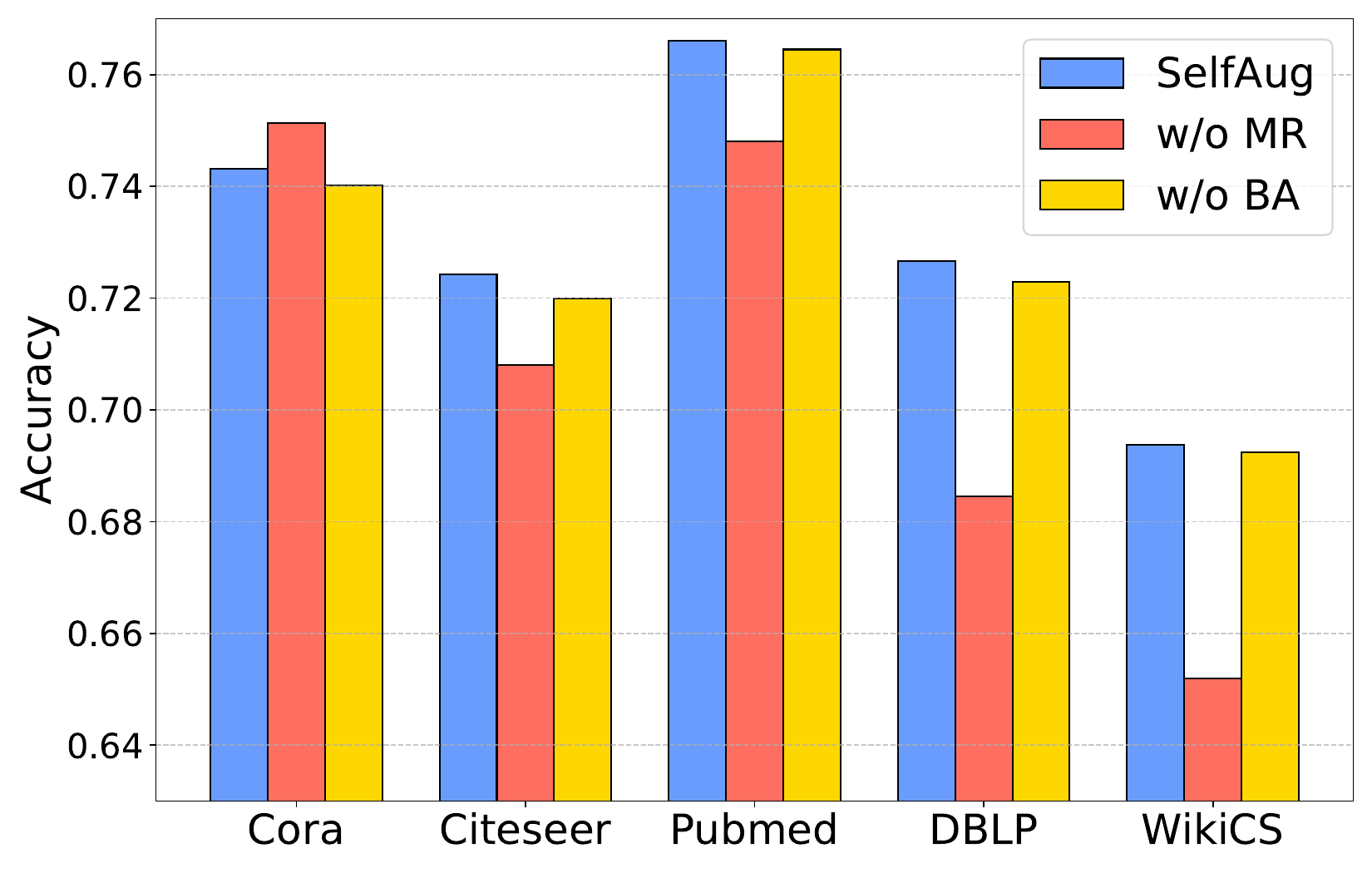}
\caption{Ablation study of Message Regularization (MR) and Bootstrap Augmentation (BA). Experiments conducted under the cold-start setting.}
\label{fig:ablation}
\end{figure}

To assess the contributions of key components in our framework, we conduct an ablation study by removing (1) the \textbf{Message Regularization} (MR) module and (2) the \textbf{Bootstrap Augmentation} (BA) strategy.
Figure~\ref{fig:ablation} reports the test accuracy across five datasets.
Removing either component leads to performance drops across most cases, highlighting their complementary roles. Specifically, removing MR causes larger performance degradation, suggesting that MR is crucial for promoting stable and generalizable neighborhood aggregation. Meanwhile, disabling the bootstrap mechanism results in smaller but still consistent declines, indicating that BA helps enhance learning under sparse supervision.

\section{Conclusion}
In this work, we proposed a novel feature-centric graph data augmentation framework that bypasses explicit structure editing by directly enhancing node embeddings through a generative, self-supervised objective. Our method learns to compensate for missing structural information by modeling embedding differences between masked and observed graphs, enabling it to generalize effectively in inductive and cold-start settings. Extensive experiments across multiple benchmarks demonstrate that our approach achieves state-of-the-art performance in both accuracy and efficiency, underscoring its potential as a scalable and generalizable solution for real-world graph learning tasks.

Despite these promising results, our current evaluation is limited to within-domain scenarios, where the training and test graphs share similar structural and feature distributions. In real-world applications, however, graphs from different domains—such as biomedical, social, or citation networks—can vary significantly in sparsity, scale, and connection patterns, posing major challenges for generalization. A key direction for future work is to explore cross-domain transfer and develop more adaptive augmentation strategies that can handle such distributional shifts, further extending the applicability of our framework to diverse and heterogeneous graph settings.

\section{Acknowledgments}
Yu Song, Bingheng Li, Jingzhe Liu, Jiliang Tang and Hui Liu are supported by the National Science Foundation (NSF) under grant numbers CNS2321416, IIS2212032, IIS2212144, IIS 2504089, DUE2234015, CNS2246050, DRL2405483 and IOS2035472, the Michigan Department of Agriculture and Rural Development, US Dept of Commerce, Gates Foundation, Amazon Faculty Award, Meta, NVIDIA, Microsoft and SNAP.

\end{document}